\documentclass[acmsmall]{acmart}

\usepackage{color}

\AtBeginDocument{%
  \providecommand\BibTeX{{%
    \normalfont B\kern-0.5em{\scshape i\kern-0.25em b}\kern-0.8em\TeX}}}
\setcopyright{acmcopyright}
\copyrightyear{2021}
\acmYear{2021}
\acmDOI{10.1145/1122445.1122456}

\acmJournal{JACM}
\acmVolume{37}
\acmNumber{4}
\acmArticle{111}
\acmMonth{7}

\begin{document}

\title{Search By Image: Deeply Exploring Beneficial Features for Beauty Product Retrieval}

\author{Mingqiang Wei}
\affiliation{%
  \institution{Nanjing University of Aeronautics and Astronautics}
  \city{Nanjing}
  \country{China}}
\email{mqwei@nuaa.edu.cn}

\author{Qian Sun}
\email{sq970306@gmail.com}
\orcid{0000-0002-5384-7411}
\affiliation{%
  \institution{Nanjing University of Aeronautics and Astronautics}
  \city{Nanjing}
  \country{China}
}

\author{Haoran Xie}
\affiliation{%
  \institution{Lingnan University}
  \city{Hong Kong SAR}
  \country{China}}
\email{hrxie@ln.edu.hk}

\author{Dong Liang}
\affiliation{%
  \institution{Nanjing University of Aeronautics and Astronautics}
  \city{Nanjing}
  \country{China}}
\email{liangdong@nuaa.edu.cn}

\author{Fu Lee Wang}
\affiliation{%
  \institution{Hong Kong Metropolitan University}
  \city{Hong Kong SAR}
  \country{China}}
\email{pwang@hkmu.edu.hk}

\renewcommand{\shortauthors}{Wei et al.}

\begin{abstract}
Searching by image is popular yet still challenging due to the extensive interference arose from i) data variations (e.g., background, pose, visual angle, brightness) of real-world captured images and ii) similar images in the query dataset.
  This paper studies a practically meaningful problem of beauty product retrieval (BPR) by neural networks. We broadly extract different types of image features, and raise an intriguing question that whether these features are beneficial to i) suppress data variations of real-world captured images, and ii) distinguish one image from others which look very similar but are intrinsically different beauty products in the dataset, therefore leading to an enhanced capability of BPR. To answer it, we present a novel \textbf{v}ariable-attention neural network to understand the combination of \textbf{m}ultiple features (termed VM-Net) of beauty product images. Considering that there are few publicly released training datasets for BPR, we establish a new dataset with more than one million images classified into more than 20K categories to improve both the generalization and anti-interference abilities of VM-Net and other methods. We verify the performance of VM-Net and its competitors on the benchmark dataset Perfect-500K, where VM-Net shows clear improvements over the competitors in terms of $MAP@7$. The source code and dataset will be released upon publication.
\end{abstract}

\ccsdesc[500]{Computing methodologies~Visual content-based indexing and retrieval}

\keywords{Beauty product retrieval, VM-Net, Multiple-feature fusion, Variable attention}

\maketitle

\section{Introduction}
Image retrieval is divided into text-based retrieval (TBR) and content-based retrieval (CBR). TBR requires professional text information, while CBR harnesses images/photos captured in real world as input. 
As an important part of CBR, beauty product retrieval (BPR) becomes more and more popular. Extensive e-commerce platforms such as Amazon, Alibaba, eBay support to \textit{search by image} for BPR. An increasing number of consumers prefer shopping online and searching for product information by capturing images from real-world products of interest anytime and anywhere. 

\begin{figure}[t]
	\centering
	\includegraphics[width=1\linewidth]{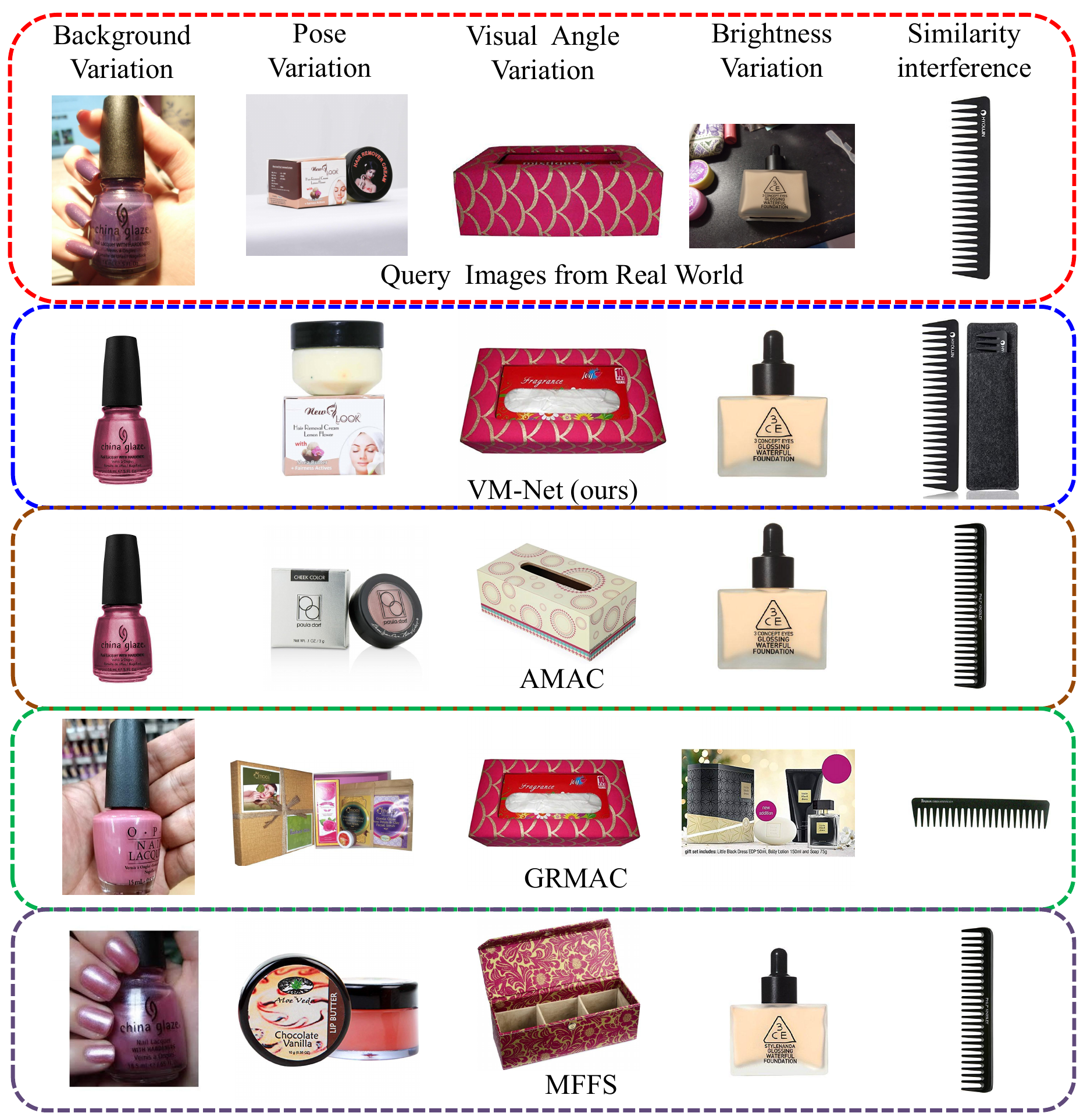}
	%
	\parbox[t]{.9\columnwidth}{
	}
	\caption{\label{fig:firstExample}
	Given a query image, finding an image containing exactly the same product in related datasets is non-trivial due to data variations ($e.g.$, background, pose, visual angle, brightness) of real-world captured images, and the extensive interference of similar images. VM-Net can flexibly deal with these harsh scenarios and outperform its competitors including AMAC \cite{DBLP:conf/mm/YuXLXHGS20}, GRMAC \cite{2019Beauty} and MFFS \cite{DBLP:conf/mm/YanLDLX20} in most cases. 
		}
\end{figure}

Despite the great improvements,
BPR is still a very challenging problem, particularly in many harsh situations~\cite{2019Beautyi,2019The,2020Multi,2019Beauty}, due to two main reasons: i) query images are often captured from real-world objects, which are full of data variations arose by background, pose, visual angle, and brightness; ii) there are a large amount of similar but intrinsically different items in a large pool of images, which are difficult to distinguish. These two types of unfavorable factors commonly lead most state-of-the-arts to the failure of BPR, as shown in Figure \ref{fig:firstExample}. 
Each year, there are many contests of BPR to enhance the retrieval accuracy, e.g., Product Identification Competition held by Alibaba~\cite{taobao}.
To the best of our knowledge, few existing algorithms can serve as a BPR panacea for various practical applications.

 In  this  paper, we propose to broadly extract different types of image features, including salient features, middle-layer features, distinct features and detailed features, to comprehensively tackle the challenges of BPR. 
 Deeply exploring these features in a neural network not only is intellectually interesting, but also enables many practical retrieval applications. To improve the capability of BPR, we aim to tackle a difficult problem of 
\begin{itemize}
  \item distinguishing very similar but intrinsically different interference images;
  \item suppressing data variations of real-world captured images.
\end{itemize}
   
Convolution-feature maps extracted from a neural network naturally contain both beneficial and interference information. 
In order to improve the retrieval accuracy, one may exploit more beneficial information while reducing interference information as much as possible.
First, to deeply exploit beneficial information, we reform Generalized-attention Regional Maximum Activation of Convolutions (GRMAC)~\cite{2019Beauty} by combining Lp-pooling, Avg-pooling and Max-pooling in an effective way, which extracts both distinct features and detailed features. 
Second, to reduce interference information, we introduce a \textit{variable-attention} mechanism (a variable mask) into Maximum Activation of Convolutions (MAC)~\cite{2016RMAC} and GRMAC which can filter background interference; we also make a saliency mask out of the salient-feature map which is close to the outline of the image to filter convolution feature maps after the variable mask, to avoid the interference of positions and directions of objects. 
By the two kinds of masks, our method can effectively eliminate the negative influence of data variations.
Based on the above improvements, we invent Generalized-attention Regional Maximum and \textit{Average} Activation of Convolutions (GRMAAC) and variable-attention based Maximum Activation of Convolutions (VAMAC).
Third, the scale of last convolution layer of a neural network is very small which inevitably loses some beneficial information. We have to consider the features of a middle-convolution layer in the backbone network as a complement to the final features.
Fourth, training the backbone network on an effective dataset not only helps enhance the ability to resist interference information, but also enhances the generalization ability. 
We establish a new dataset which contains more than one million images classified into more than 20K categories to train our backbone network. 
%
%

The core contributions of our work are summarized as follows:
\begin{itemize}
	\item We propose an end-to-end variable-attention and multi-feature-combination based neural network (VM-Net) to solve the practical yet challenging BPR problem.
	\item We propose GRMAAC and VAMAC by reforming the pooling solution of GRMAC and introducing the variable-attention mechanism into GRMAC and MAC.
	\item We make full use of four types of beneficial features including salient features, middle-layer features, distinct features and detailed features, to improve the performance of VM-Net.
	\item We collect a large dataset and train the backbone network on this dataset by our parameter adjustment scheme, ensuring the network's effectiveness on BPR.
\end{itemize}
We conduct extensive experiments on the Perfect-500K dataset which contains more than half million beauty products.  Perfect-500K is considered as the publicly-released most challenging dataset on the BPR task. The results show clear improvements of VM-Net over its competitors in terms of retrieval accuracy.

\section{Related Work}
\textbf{Content-based image retrieval (CBR)}. CBR aims to find an exact image from a given database, which contains the same instance as the given query image. Since 2003, image retrieval based on local descriptors and global descriptors has been extensively studied for over a decade~\cite{DBLP:conf/iccv/BrattoliRO19,DBLP:conf/cvpr/JangC20,DBLP:journals/csur/DattaJLW08,2018SIFT}. Recently, image representations based on the convolutional neural network (CNN) have attracted increasing interest in the community and demonstrated impressive performance. For example, Revaud et al.~\cite{DBLP:conf/iccv/RevaudARS19} train the retrieval network with a List-wise Loss to optimize the global ranking and consider a large number of images concurrently. Sun et al.~\cite{DBLP:conf/cvpr/SunCZZZWW20} introduce Circle Loss into a retrieval network, aiming to maximize the within-class similarity and minimize the between-class similarity. Overall, in the last decade, the image retrieval task has become a greatly developed and highly-focused task in academia and industry~\cite{2007A,DBLP:conf/eccv/XuanSLP20,DBLP:conf/eccv/BrownXKZ20,DBLP:conf/eccv/GeW00L20}. Particularly, beauty product retrieval (BPR) has gained more and more attention for its wide application prospects~\cite{2019Beautyi,2019The,2019Beauty,2019Beaut}.

\textbf{GRMAC}. Yu et al.~\cite{2019Beauty} propose the Generalized-attention Regional Maximum Activation of Convolution (GRMAC) that improves the accuracy of BPR. The method introduces a weight matrix into the Regional Maximum Activation of Convolutions (RMAC)~\cite{2016RMAC}, where more features related to the content can be extracted and the background can be suppressed at the same time. Although the method improves the retrieval accuracy to a certain degree, it pays much attention to extracting distinct features, while detailed features which contain significant information of contents are ignored. Besides, there is room for improving the ability of GRMAC to remove the background interference. We attempt to make full use of both detailed features and distinct features to improve the retrieval capability of GRMAC.

\textbf{Attention mechanism}. In the field of image retrieval, especially BPR, backgrounds and data variations bring big challenges in the accuracy of retrieval results. Integrating attention mechanism into global descriptors and local descriptors may alleviate these problems. For example, Ng et al.~\cite{DBLP:conf/eccv/NgBTM20} combine attention mechanism and second-order loss to train the backbone network of image retrieval. Fang et al.~\cite{DBLP:conf/iccv/FangZRPH19} propose bilinear attention networks (BAN) by applying bilinear attention to CNN, which enhances the connection between bilinear attention and given vision information. Yu et al.~\cite{DBLP:conf/mm/YuXLXHGS20} propose Attention-based MAC (AMAC) which combines an invariant mask with MAC to reduce the interference of interference information. It is necessary to take different attention masks according to the change of the characteristics of tasks, while the traditional invariable-attention mechanism cannot meet the demand.

\textbf{Low-or-medium-level features}. There are many effective low-level or medium-level features which contain rich information of images, such as edge map, saliency map, middle-convolution-layer features and so on. In the past cases of combining CNN and image retrieval, researchers focused on high-level features of CNN while ignored the effect of low-level features and medium-level features. Wei et al.~\cite{DBLP:conf/cvpr/WeiWWSH020} propose Label Decoupling Framework (LDF) for salient object detection. Due to the outstanding anti-interference ability of saliency map, we introduce LDF into our method to extract the saliency map and use it to improve the performance of neural network on resisting the interference of data variants. Redmon et al.~\cite{DBLP:journals/corr/abs-1804-02767} introduce Feature Pyramid Networks (FPN) into Yolo-v2~\cite{DBLP:conf/cvpr/RedmonF17} to capture the characteristics of objects with different sizes for object detection. In the field of BPR, there is a similar problem that the highest layer of CNN inevitably loses some useful information, so it is necessary to introduce FPN into our method.

\section{Method}
By resorting to dual measures simultaneously, we exploit beneficial features while suppressing interference features in a well-designed neural network that improves the accuracy of BPR as much as possible. Therefore, the first vital thing is to reform existing convolutions to exploit the beneficial distinct features and detailed features; the second vital thing is to introduce effective mechanisms (e.g., attention) into existing convolutions to suppress interference features. In addition, both the backbone network and its pretraining should be considered to further improve the accuracy of BPR, thus, training an appropriate backbone network on an appropriate  dataset not only helps enhance the ability to resist interference information, but also enhances the generalization ability.

\begin{figure}[htb]
	\centering
	\includegraphics[width=1\linewidth]{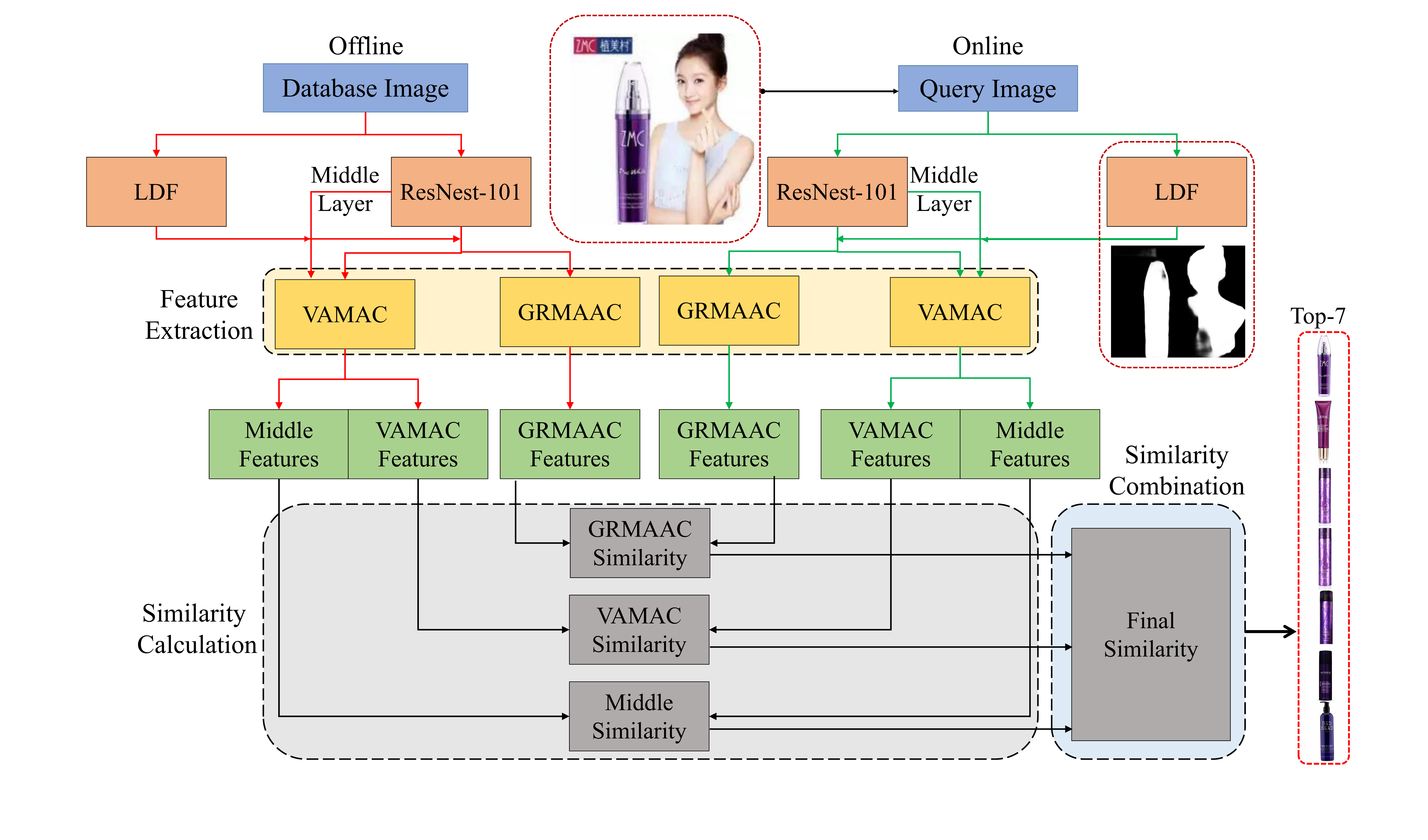}
	%
	\parbox[t]{.9\columnwidth}{
	}
	\caption{\label{fig:pipeline}
	Architecture of VM-Net.  We extract the features of database images which are divided into GRMAAC features, VAMAC features and Middle features, and save them as database features in an offline way. Given a query image, we extract its features in the same way, and calculate the similarity of each feature respectively and summarize these three similarities. Finally, the query image is compared  with each database image by the final similarity, and seven most similar images are obtained as the online retrieval results. The green line represents the process of online extraction of the query image's features. The black line represents the process of online calculation and combination of similarities. The red line represents the process of offline extraction of database images' features.}
\end{figure}

Based on the aforementioned observations, we propose an end-to-end variable-attention and multi-feature-combination based neural network (VM-Net) to solve the practical yet challenging BPR problem. The proposed VM-Net is illustrated in Figure \ref{fig:pipeline}: 
(1) We utilize the effective ResNest-101 which is pretrained on ImageNet~\cite{DBLP:conf/cvpr/DengDSLL009} as the backbone network and train it on the newly constructed dataset (a dataset that contains more than one million images classified into more than 20K categories). (2) We design both GRMAAC (mainly for exploiting distinct features and detailed features and fusing them) and VAMAC (mainly for suppressing interference features) as new descriptors, which are supported by the saliency masks extracted by label decoupling framework (LDF)~\cite{DBLP:conf/cvpr/WeiWWSH020}. (3) As known, the scale of last convolution layer of a neural network is very small which inevitably loses beneficial information. We have to consider the features of a middle-convolution layer in the backbone network as a complement to the final features. We take advantage of feature pyramid networks (FPN) \cite{cvpr/LinDGHHB17} to extract middle-convolution-layer feature maps. (4) Totally, the above three kinds of features are extracted from both the query image and database images, namely GRMAAC features, VAMAC features, Middle features. Afterwards, we calculate the Middle similarity, the VAMAC similarity and the GRMAAC similarity between the query image and each image in the database, and the final similarity is made up of these three similarities. The top-seven images are sorted as the retrieval results based on the final similarity. It is worth noting that, the feature extraction of database images is in an offline way, and the feature extraction of a query image and calculation and combination of similarities are in an online way.

\begin{figure}[t]
	\centering
	\includegraphics[width=0.95\linewidth]{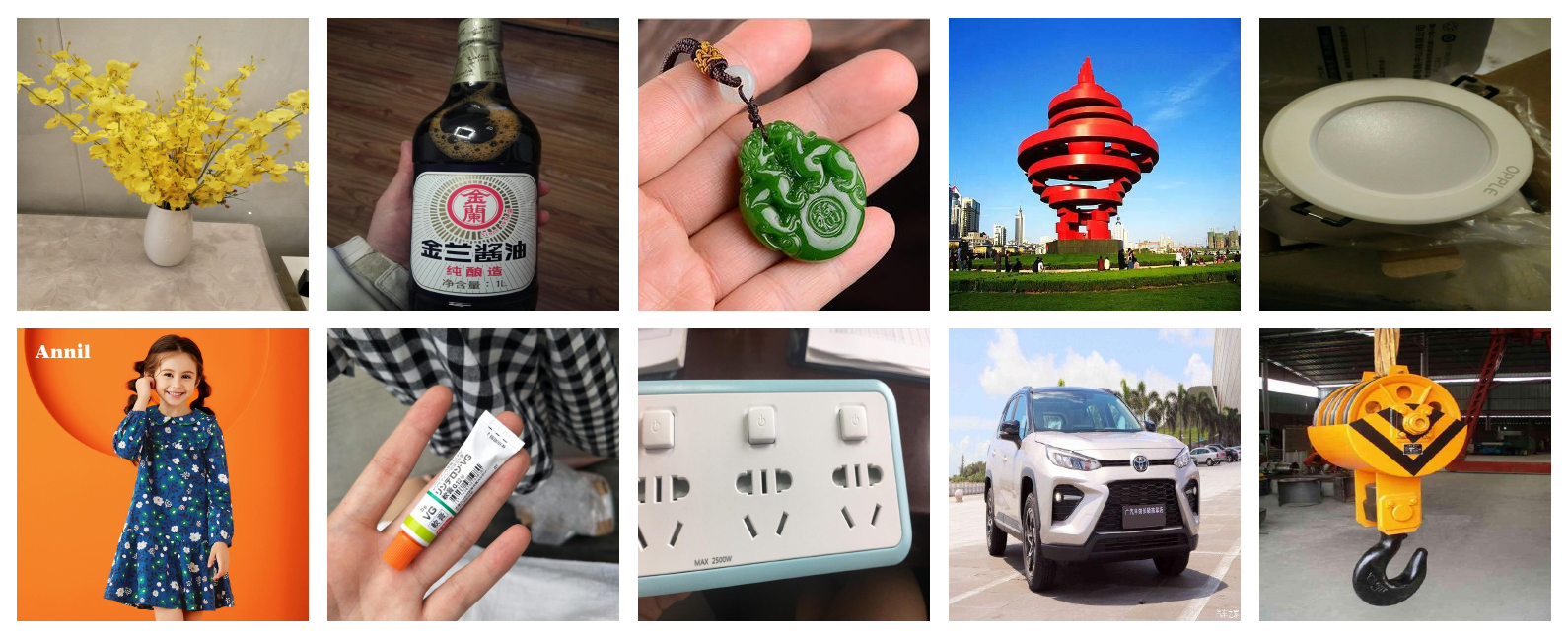}
	%
	\parbox[t]{.9\columnwidth}{
	}
	\caption{\label{fig:traindataset}
	Some images from our established dataset. 
		}
\end{figure}
\begin{figure}[htb]
	\centering
	\includegraphics[width=0.95\linewidth]{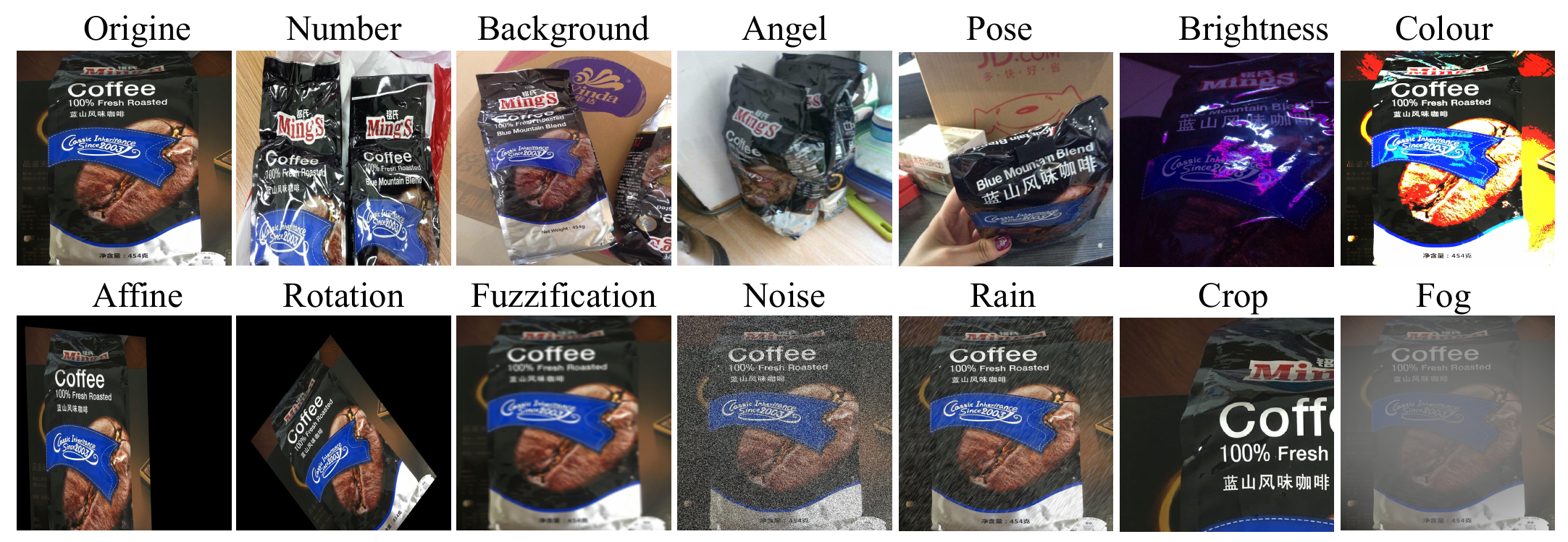}
	%
	\parbox[t]{.9\columnwidth}{
	}
	\caption{\label{fig:augmentation}
Data augmentation on an image from our dataset. There is relatively large interference information in the image that the training process can improve the generalization ability and the robustness of the backbone network to deal with jamming information. }
\end{figure}

\subsection{Backbone Network and Training}
Backbone network has a relatively large impact on both the accuracy and speed of learning-based methods. We select ResNest-101 as the backbone network of VM-Net by simultaneously considering its size, running speed and performance.
We pretrain ResNest-101 on ImageNet, and again train it on our own constructed dataset. 
The biggest difficulty of BPR is that query images are often captured from real-world objects, which are full of distracting information, as shown in Figure \ref{fig:firstExample} and the number of database images is very huge, which challenges the generalization ability of the algorithm. Therefore, it is necessary to improve the generalization ability and the ability to counter jamming information of a retrieval method. Based on such demands, we create a new dataset from two main aspects. First, as shown in Figure \ref{fig:traindataset}, we collect more than 20K kinds of images from various fields such as machinery, medicine, buildings, e-commerce, etc. Second, as shown in Figure \ref{fig:augmentation}, we augment the data including geometric transformation and increasing interference, such as background variation, brightness variation, noise, blurring, etc. Please note that in order to improve the versatility of the dataset and enhance the capability of the model in other image retrieval tasks, we cover many kinds of images, which are not limited to beauty product images.

\subsection{LDF-based VAMAC}
Maximum Activation of Convolutions (MAC) is a common feature descriptor in the field of CBR which can effectively extract the capital features of images. MAC is formulated as
$$f_{MAC} = [f_1, ..., f_i..., f_C],  f_i = \max\limits_{x \in X_i} x \eqno{(1)}$$
where $X_i$ represents the convolutional feature map extracted by the backbone network belonging to the i-th channel, and $f_i$ expresses the biggest element of the i-th feature map.

\begin{figure}[htb]
	\centering
	\includegraphics[width=0.95\linewidth]{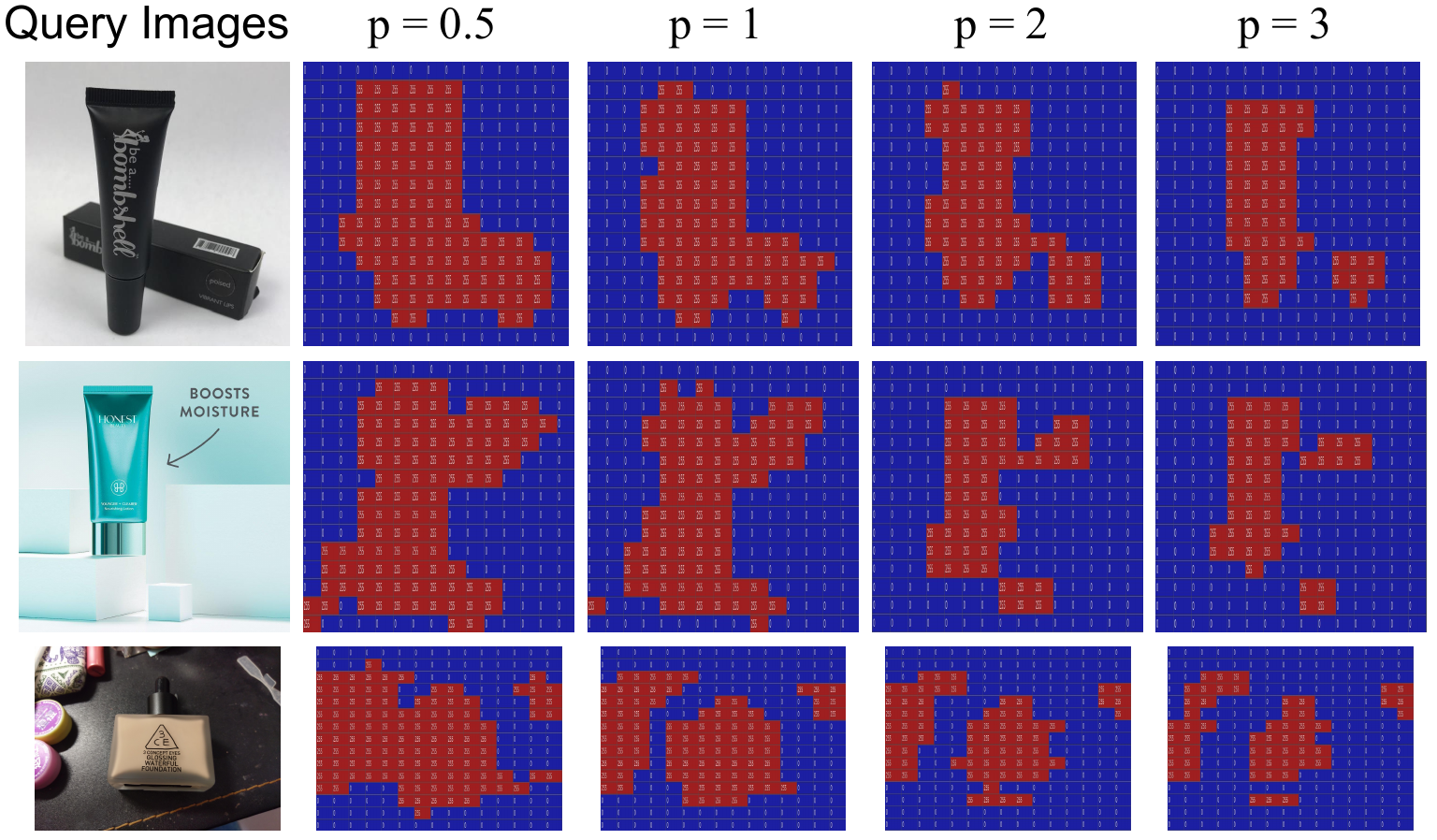}
	%
	\parbox[t]{.9\columnwidth}{
	}
	\caption{\label{fig:mask}
		Visualization of variable masks. The variable mask is controlled by the parameter $p$. As $p$ gets bigger, the scope of the mask gets smaller.  }
\end{figure}

Unfortunately, images of BPR are full of distracting information which may be more visible than the content of the image and interfere with the feature extraction of MAC. To reduce this interference, we introduce a variable-attention mechanism into MAC, i.e., propose variable-attention based maximum activation of convolutions (VAMAC). The variable-attention mechanism is inspired by Generalized-attention Regional Maximum Activation of Convolutions (GRMAC), which uses the feature maps of the last convolution-layer of the backbone network to generate a variable weight matrix that is only used to assign weights to sliding windows. According to its principle that we believe that areas with element values bigger than the average contain products, we generate a variable mask to filter feature maps in order to reduce interference information. The variable mask $M$ is generated by defining a threshold $T$ (the average value) as
$$A = \sum_{i=1}^Cf_i \eqno{(2)}$$
$$minA = \min\limits_{x \in A} x \eqno{(3)}$$
$$T = (\frac{1}{w \times h}\sum_{i=1}^w\sum_{j=1}^h(A_{i,j} - minA)^p)^{\frac{1}{p}} + minA \eqno{(4)}$$
$$ M_{i,j}=\left\{
\begin{array}{rcl}
1      &      & {A_{i,j} > T}\\
0      &      & {otherwise}
\end{array} \right. \eqno{(5)}$$
where $A$ denotes the summation of feature maps through depth. $T$ is a threshold that controls the generation of a variable mask. Values of elements bigger than $T$ in $A$ are set to 1 and the others are set to 0, the variable mask $M$ is generated. $p$ is a parameter to tune for adjusting the variable mask, as shown in Figure \ref{fig:mask}. VAMAC filters feature maps with the variable mask before MAC defined as 
$$y_{i,j} = M_{i,j} \times x_{i,j} \eqno{(6)}$$
$$f_{VAMAC} = [f_1...f_c...f_C], with f_c = \max\limits_{y \in Y_c} y \eqno{(7)}$$
where $M_{i,j}$ is the value of variable mask in position (i,j) as shown in equation (5) and $x_{i,j}$ is the value of feature map in position (i,j). In addition to the advantage of reducing interference information, VAMAC is also adjustable compared with other attention mechanisms and this advantage ensures that it can adjust its variable mask as the change of task's characteristics, so as to better capture useful information.

\begin{figure}[htb]
	\centering
	\includegraphics[width=0.95\linewidth]{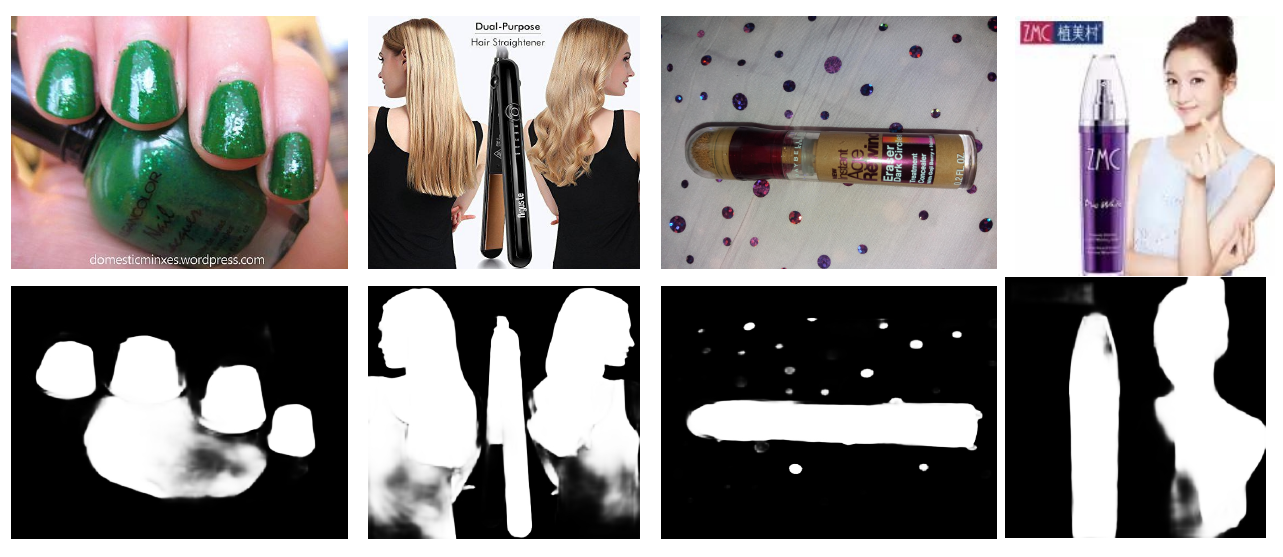}
	%
	\parbox[t]{.9\columnwidth}{
	}
	\caption{\label{fig:saliency}
		Visualization of saliency maps. LDF~\cite{DBLP:conf/cvpr/WeiWWSH020} can extract saliency maps which are very close to the outline of contents. }
\end{figure}

Although VAMAC alleviates the influence of interference information, it cannot remove the interference information near the object contour and it is difficult to fight the interference of positions and directions. In order to remedy these shortcomings, we use label decoupling framework (LDF) to extract the saliency map which is a kind of low-level features and is very close to the outline of the content. It can remove the interference information near the object contour and fight the interference of positions and directions well, as shown in Figure \ref{fig:saliency}. Then, we subsample the saliency map with bilinear interpolation to the same scale as feature maps and use it to filter feature maps which have been filtered by the above variable mask. 


\subsection{LDF-based GRMAAC}
\begin{figure}[htb]
		\centering
		\includegraphics[width=1\linewidth]{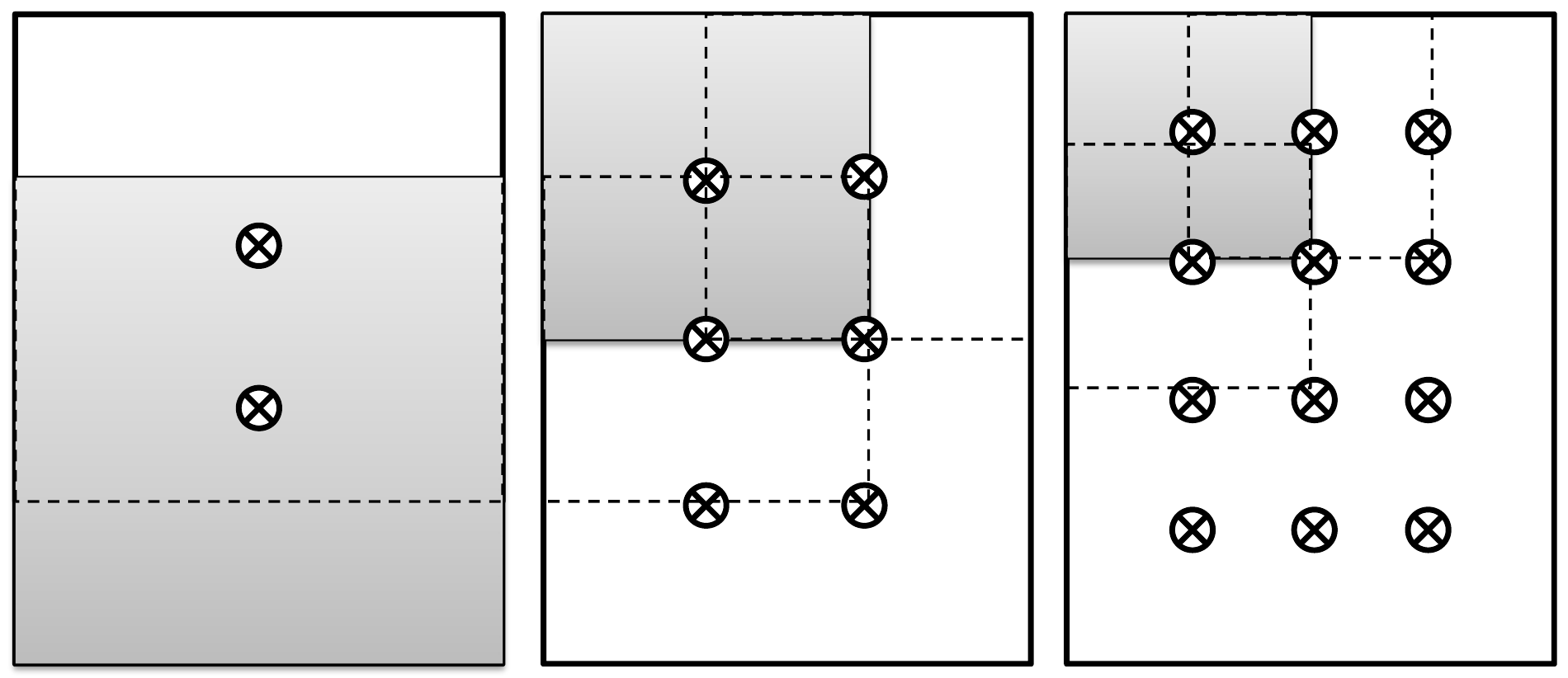}
		%
		\parbox[t]{.9\columnwidth}{
		}
		\caption{\label{fig:square}
			Visualization of sliding windows from GRMAC. GRMAC uses three scales of sliding windows to extract local features within regions of different scales. }
	\end{figure}

GRMAC is a famous descriptor for sliding windows and the weight matrix in the field of BPR. As shown in Figure \ref{fig:square}, sliding windows have three different scales and they can extract local features. GRMAC applies Max-pooling on all scales of sliding windows to capture most distinct features while ignoring the detailed ones. There are many images in the database that are different from but similar to the query image, which may cause extreme interference to the retrieval accuracy. Detailed features are sensitive to the nuances of similar images. In this regard, we propose GRMAAC by introducing multi-pooling into GRMAC to extract both detailed features and distinct features, and then fuse them.
We find that the larger window is suitable for extracting more detailed features since it contains more areas, while the smaller window is suitable for capturing distinct features within a small area. Meanwhile, Avg-pooling can capture more detailed features and Max-pooling is easier to extract more distinct features. Therefore, we assign Avg-pooling to the biggest window, $L_{p}$-pooling to the medium window, and Max-pooling to the smallest window. $L_{p}$-pooling is a pooling solution between Max-pooling and Avg-pooling, and we can adjust its parameters to decide whether to  extract more detailed features or catch more distinct features.
In order to compute them, we obtain three set of square regions sampled by the sliding windows: $R_1$ = \{$R_{k,1}$\}, $R_2$ = \{$R_{k,2}$\} and $R_3$ = \{$R_{k,3}$\}. $R_1$ are the largest ones, $R_3$ are the smallest ones and $R_2$ are the medium ones. GRMAAC is calculated as
$$f_{R_{k,1}} = [f_{R_{k,1,0}}...f_{R_{k,1,c}}...f_{R_{k,1,C}}], f_{R_{k,1,c}} = \max\limits_{x \in R_{k,1,c}} x \eqno{(8)}$$
$$f_{R_{k,2}} = [f_{R_{k,2,0}}...f_{R_{k,2,c}}...f_{R_{k,2,C}}], f_{R_{k,2,c}} = lp_{x \in R_{k,2,c}} x \eqno{(9)}$$
$$f_{R_{k,3}} = [f_{R_{k,3,0}}...f_{R_{k,3,c}}...f_{R_{k,3,C}}], f_{R_{k,3,c}} = avg_{x \in R_{k,3,c}} x \eqno{(10)}$$
$$f_{GRM} = \sum_{t \in {1,2,3}}\sum_{R_{k,t} \in R_t}q_t\frac{\sum_{(i,j) \in r_{k,t}} M_{i,j}}{size(R_{k,t})}f_{R_{k,t}}\eqno{(11)}$$
$$f_{GRMAAC} = \frac{f_{GRM}}{norm(f_{GRM})} \eqno{(12)}$$
where $f_{R_{k,t}}$ denotes the local feature of $R_{k,t}$. $r_{k,t}$ represents the location set of $R_{k,t}$. $M_{i,j}$ represents the value of mask in location (i,j). $q_t$ represent the weights which we assign to different pooling schemes and they are adjustable. Therefore, we can fuse detailed features and distinct features by adjusting $q_t$ to appropriate value.

Similar to VAMAC, GRMAAC also needs LDF and variable mask to enhance the ability of anti-jamming information. To the Avg-pooling part, we use the variable mask and the saliency mask to filter the feature maps before Avg-pooling, because the sliding windows in this part are big and have much distracting and interference information which may easily affect Avg-pooling. As for the $L_{p}$-pooling and Max-pooling, we only use the variable mask to filter the feature maps' distracting information for the reason that the sliding windows in this part are relatively small so that they contain less interference information and the two pooling methods are more difficultly affected by interference information relatively.

\subsection{Middle Features}
In the past cases of combining CNN and image retrieval, researchers only focused on the last convolution-layer feature maps while they ignored to make use of the middle-layer features. Although, the last convolution-layer feature maps contain most useful information, they will inevitably lose some useful information due to the limitation of size. Obviously, it is a good choice to use the middle-layer feature maps as supplementary features.

We introduce feature pyramid networks (FPN) into our method to extract the middle-layer feature maps of the backbone network. First, we install FPN on the backbone network and freeze all parts except FPN. Second, we pretrain the network on Imagenet and train it on our own dataset mentioned above. After that, FPN can extract the middle-layer feature maps while the backbone network is working. It is worth mentioning that FPN is only used to extract middle-layer feature maps, and has no influence on last-layer feature maps. Then, after applying VAMAC and LDF on the middle-layer feature maps, the middle features are obtained.
\subsection{VM-Net}

We propose VM-Net by combining VAMAC features, GRMAAC features and Middle features in the stage of calculating similarity of the database images and the query image. In order to balance accuracy and speed, we use cosine similarity to calculate the similarity of feature vectors. Feature vectors have been normalized before, so it is not necessary to divide the length of the vector when calculating the cosine similarity. We first calculate their contributions to similarity separately, i.e., GRMAAC similarity, VAMAC similarity and Middle similarity. Then, we combine them with the weights $p_{s,t}$ according to their importance. VM-Net is formulated as
$$S_{VA} = f_{VA,query} \cdot f_{VA,db}\eqno{(13)}$$
$$S_{GR} = f_{GR,query} \cdot f_{GR,db}\eqno{(14)}$$
$$S_{M} = f_{M,query} \cdot f_{M,db}\eqno{(15)}$$
$$S = p_{s,1} \times S_{VA} + p_{s,2} \times S_{GR} + p_{s,3} \times S_{M}\eqno{(16)}$$
where, $f_{VA,query}$ and $f_{VA,db}$ denote VAMAC features of the query image and the database image. $f_{GR,query}$ and $f_{GR,db}$ denote GRMAAC features of the query image and the database image. $f_{M,query}$ and $f_{M,db}$ denote Middle features of the query image and the database image. $S_{VA}$, $S_{GR}$ and $S_{M}$ represent VAMAC similarity, GRMAAC similarity and Middle similarity respectively. Finally, the similarity of the database images and the query image is generated by combining VAMAC similarity, GRMAAC similarity and Middle similarity with weights $p_{s,t}$.

\subsection{Usage of VM-Net}
As shown in Figure \ref{fig:pipeline}, our VM-Net is divided into two main parts, including extracting database features offline and extracting the feature set of the query image and comparing it with database features online. In the offline stage, we extract the features of images from the database and save them as a database feature set. In the online stage, users can get the top seven most similar images after our method extracts the features of the query image and compares it with each feature group from the database feature set.
\subsection{Innovations}
In this section, we shall elaborate our innovations.
\begin{itemize}
	\item We establish a new and large dataset which enhances the generalization ability and anti-interference ability of the proposed VM-Net and other methods by training the backbone networks (i.e., ResNest-101) on it.
	\item We introduce LDF and FPN into VM-Net to make use of low-level features (Saliency mask) and middle-convolution-layer features (Middle features). The Saliency mask helps remove the interference information near the object contour and fights the interference of position and direction well. Middle features make up for the disappearance of useful information in the last-convolution-layer feature maps. 
	\item We propose GRMAAC and VAMAC by introducing variable-attention mechanism into GRMAC and MAC and reforming the pooling solution of GRMAC. These improvements encourage VAMAC filter interference information flexibly and give GRMAAC an ability to capture both detailed features and distinct features simultaneously.
	\item VM-Net clearly outperforms the state-of-the-art methods in terms of retrieval accuracy. 
\end{itemize}
\section{Experiments}
We implement our VM-Net with PyTorch on a GeForce RTX 3090 GPU.
We carry out three types of experiments to demonstrate the superiority of the proposed VM-Net. First, we elaborate training details of the backbone network. Second, we evaluate the influence of all parts of VM-Net to prove their effectiveness. Third, we compare VM-Net with the state-of-the-art methods. Experiments are conducted on the Perfect-500K dataset \cite{acmchallenge}. Perfect-500K contains more than half million images of beauty/personal products collected from online shopping websites. There are 100 images for testing. The image retrieval task requires the algorithm to be able to retrieve all positive examples and arrange them in the first few examples. Mean average precision@N (MAP@N) is interested in the number and ranking of correct products in the retrieval result, where with the increase of N, the accuracy of the result is more difficult to increase. Mean average precision@7 (MAP@7) can measure this requirement better. Thus, we adopt MAP@7 as the metric.
\subsection{Evaluation Metric}
In our experiments, we adopt mean average precision@7 (MAP@7) as the evaluation metric and we choose seven images with the highest scores as the retrieval results.

MAP@k not only requires that the correct goods are retrieved as much as possible in the search results, but also requires that these correct goods be ranked as high as possible in the search results, taking into account the requirements of recall and precision. However, neither recall nor accuracy can verify the ranking of correct products in the results. When retrieving goods, users not only need to retrieve the correct goods, but also want the correct goods to be presented as soon as possible, without the need to search through the drop-down menu. The commodity retrieval algorithm with a high MAP@k can undoubtedly improve the retrieving experience of users.

MAP@7 is defined as
$$AP(n,j) = \frac{\sum_{i=1}^n(rel(i)\times{\frac{rank \quad in \quad relevant \quad images}{i}})}{number \quad of \quad relevant \quad images'}\eqno{(16)}$$
$$MAP(n) = \frac{\sum_{j=1}^{100}AP(n,j)}{100}\eqno{(17)}$$
$$MAP@7 = \frac{\sum_{n=1}^{7}MAP(n)}{7}\eqno{(18)}$$
where $j$ denotes the order of the query image and rel(i) is equal to 1 when the selected image is relevant to the query image.

\subsection{Backbone Network}
Choosing a fast and effective neural network as the backbone network is more desirable for the BPR task. As shown in Table \ref{backbone network}, ResNest-101 is the best choice by taking the accuracy rate of image classification task as the main standard, as well as considering all other indexes.

The training process of our backbone network is divided into two parts. First, we train the backbone network on ImageNet. Then, we further optimize the backbone network on the newly established dataset. It is worth mentioning that we use Softmax Loss~\cite{DBLP:conf/emnlp/AuliL11} as the loss function. The hyper-parameters are shown in Table \ref{train parameter}. In order to make the training process more smooth, we use Cosine-Decay to adjust the learning rate. To reduce the risk of over-fitting, we choose SGD as the parameter optimizer.

\begin{table}[h] 
\footnotesize
	\caption{Retrieval performance of different backbone networks. } 	\centering
	\begin{tabular}{ c c c c}
		\hline
		    & SIZE & GFLOPs & MAP@7 \\
		\hline
		ResNet-101   &    44.5M &   7.87 &0.5700\\
		
		ResNeXt-101    &  44.3M &   7.99 &0.5874\\
		
		SENet-101    &    49.2M &    8.00 &0.5808\\
		
		ResNetD-101    &   44.6M &   8.06 &0.5965\\
		
		SKNet-101      &    48.9M &   8.46&0.5950\\
		
		ResNest-101    &     48.2M &8.07 &\textbf{0.6061}\\
		VGG-16  &   138.4M & 15.5 &0.4962\\
		RepVGG-B2g4 &55.8M & 11.3 &0.5634\\
		EfficientNet-B0 &5.26M &0.4 &0.5050\\
		
		\hline
	\end{tabular} \label{backbone network}
\end{table}

\begin{table}[h] 	
	\caption{Hyper-parameters of training. } 	\centering
	\footnotesize
	\begin{tabular}{ c c c c c c c c c }
		\hline
		Parameter & batchsize & num\_epochs & epoch\_decay & lr\_decay & lr\_init & lr\_final & momentum & weight\_decay\\
		\hline
		Value & 32 & 80 & 20 & 0.1 & 0.1 & 0.0001 & 0.9 & 0.0001\\
		
		\hline
	\end{tabular} \label{train parameter}
\end{table}
\begin{table}[htp] 	
\footnotesize
	\caption{Influence of individual parts of VM-Net. MF means middle features.} 	\centering
	\begin{tabular}{ c c c c c }
		\hline
		GRMAAC& VAMAC & Saliency & MF & MAP@7 \\
		\hline
		$\surd$   &     &          & &0.5324\\
		
		&  $\surd$ &            & &0.5361\\
		
		$\surd$    &    $\surd$ &  & &0.5706 \\
		
		$\surd$ &    &          & $\surd$& 0.5308\\
		
		&    $\surd$ &        &$\surd$ &0.5432\\
		
		$\surd$    &   $\surd$ &    &$\surd$ & 0.5795\\
		
		$\surd$ &         &         $\surd$ & &0.5675\\
		&       $\surd$ &          $\surd$ & &0.5616\\		
		$\surd$  &   $\surd$ &  $\surd$ & &0.5972\\
		
		$\surd$ &            &       $\surd$&$\surd$&0.5723 \\
		
		&       $\surd$  &          $\surd$&$\surd$ &0.5650\\
		$\surd$ &      $\surd$ &   $\surd$  & $\surd$&\textbf{0.6061}\\
		
		\hline
	\end{tabular} \label{influence}
\end{table}


\begin{table}[h] 	
	\caption{Comparisons with SOTAs. The accuracy of methods marked with "AMB" comes from the official ranking of "AI meets beauty"~\cite{acmchallenge}. } 	\centering
	\begin{tabular}{ c     c     c }
		\hline
		& Method & MAP@7 \\
		\hline
		1   &  RMAC~\cite{2016RMAC} &    0.3804 \\
		2    &   GEMLL~\cite{DBLP:conf/iccv/RevaudARS19} &    0.4017 \\
		3   &    UEL~\cite{2019Beautyi} &     0.4050 \\
		4  &   GRMAC~\cite{2019Beauty} &    0.4624 \\
		5    &    MFFS~\cite{DBLP:conf/mm/YanLDLX20} & 0.4688\\
		6   &   SOLAR~\cite{DBLP:conf/eccv/NgBTM20} &    0.4850 \\
		7    &    AMAC ~\cite{DBLP:conf/mm/YuXLXHGS20}& 0.5750\\
		8   &    VM-Net (ours)        &    \textbf{0.6061} \\
		9   &    UEL (AMB)~\cite{2019Beautyi} &     0.4072 \\
		10   &   GRMAC (AMB)~\cite{2019Beauty} &    0.4086 \\
		11    &    MFFS (AMB)~\cite{DBLP:conf/mm/YanLDLX20} & 0.4373\\
		12     &    AMAC (AMB) ~\cite{DBLP:conf/mm/YuXLXHGS20}& 0.5321\\
		13     &    FIRe~\cite{DBLP:journals/corr/abs-2201-13182}& 0.5600\\
		14     &    CVNet~\cite{DBLP:journals/corr/abs-2204-01458}& 0.5250\\
		\hline
	\end{tabular}\label{compaire}
\end{table}

\subsection{Ablation Study}
We conduct ablation studies by removing or replacing components from the full implementation of VM-Net. There are five main parts, including the backbone, Saliency mask, Middle features, GRMAAC and VAMAC. Since ResNest-101 is proved to be the best choice for the backbone network, we herein evaluate the influence of the other parts. As illustrated in Table \ref{influence}, the four parts all have different degrees of positive impacts on VM-Net. GRMAAC, VAMAC, and the saliency mask focus on detailed features, distinct features, and salient features respectively, while the middle-convolution-layer features are supplements. It is observed from Table \ref{influence} that GRMAAC, VAMAC and the saliency mask contribute more than the middle-convolution-layer features.

\begin{table}[h] 	
	\caption{\noindent Adjustment of $p$ in VAMAC. } 	\centering
	\begin{tabular}{ c c c}
		\hline
		     & $p$ & map@7 \\
		\hline
		1   &    0.5 &    0.5658 \\
		
		2    &   \textbf{1.0} &    \textbf{0.5697} \\
		
		3    &    2.0 &    0.5672 \\
		
		4    &    3.0 &    0.5614 \\
		\hline
	\end{tabular} \label{adust_p}
\end{table}
\begin{table}[h] 	
	\caption{\noindent Adjustment of $q_t$ in GRMAAC.} 	\centering
	\begin{tabular}{ c    c    c   c }
		\hline
		    & $q_1$ & $q_2$ & map@7 \\
		\hline
		1   &    \textbf{0.5} &   \textbf{0.5} & \textbf{0.5502}\\
		
		2    &   0.5 &   1.0 & 0.5495\\
		
		3    &   0.5 &    2.0 & 0.5406\\
		
		4    &   1.0 &    0.5 & 0.5469\\
		5    &   1.0 &    1.0 & 0.5425\\
		6    &   1.0 &    2.0 & 0.5357\\
		7    &   2.0 &    0.5 & 0.5313\\
		8    &   2.0 &    1.0 & 0.5229\\
		9    &   2.0 &    2.0 & 0.5176\\
		\hline
	\end{tabular} \label{adust_pt}
\end{table}
\begin{table*}[htp]
 \small	
	\caption{\noindent Adjustment of $p_{s,t}$ in VM-Net.} 	\centering
	\begin{tabular}{c c c c c c c c c c c c c}
		\hline
		& 1 & 2 & 3 & 4 & 5 & 6 & 7 & 8 & 9 & 10  \\
		\hline
		$p_{s,2}$  & 0.9 & 1.0 & 1.2 & 1.3 & 1.4 & 1.5 & 1.6 & \textbf{1.7}&1.8&1.9 \\
		map@7   & 0.5932 & 0.5995 & 0.5975 & 0.5986 & 0.6047 & 0.6030 & 0.6052 & \textbf{0.6061} & 0.6032&0.5924 \\

		\hline
	\end{tabular}\label{adust_qt}
\end{table*}

\subsection{Parameter Adjustment}
In order to obtain the best result, we adjust the parameters of VAMAC, GRMAAC and VM-Net separately by conducting extensive experiments. The way we take to adjust parameters is that we change one parameter and fix the other parameters at a certain time while observing its impact on the accuracy.

Through extensive experiments and tests, we obtain a series of optimal parameters of VM-Net. The most important three parameters are introduced as follows:
\begin{itemize}
	\item We set $p=1.0$ which controls the variable mask.
	\item We adjust $q_t$ for GRMAAC which represents the weights assigned to different pooling schemes. Explicitly, we set $q_1=0.5$, $q_2=0.5$, and $q_3=1.0$.
	\item We adjust $p_{s,t}$ which affects the VAMAC similarity, the GRMAAC similarity and the Middle similarity, respectively. We set $p_{s,1}=1.0$, $p_{s,2}=1.7$, and $p_{s,3}=1.0$.
\end{itemize}

\subsection{Comparison}
To prove the effectiveness of VM-Net, we compare it with the state-of-the-art methods including AMAC~\cite{DBLP:conf/mm/YuXLXHGS20} (the grand-challenge champion of ACM MM 2020), MFFS~\cite{DBLP:conf/mm/YanLDLX20} (the grand-challenge second place of ACM MM 2020), GRMAC~\cite{2019Beauty} (the grand-challenge champion of ACM MM 2019), UEL~\cite{2019Beautyi} (the grand-challenge second place of ACM MM 2019), SOLAR~\cite{DBLP:conf/eccv/NgBTM20} (proposed by Ng et al. in ECCV 2020), GeMLL~\cite{DBLP:conf/iccv/RevaudARS19} (proposed by Revaud et al. in ICCV 2019), FIRe~\cite{DBLP:journals/corr/abs-2201-13182}, CVNet~\cite{DBLP:journals/corr/abs-2204-01458} and RMAC~\cite{2016RMAC} (a classic image retrieval method). All these methods are tested on Perfect-500K. As shown in Table \ref{compaire}, our VM-Net significantly improves the retrieval accuracy compared with these SOTAs.
It is worth mentioning that the accuracy of UEL, GRMAC, MFFS and AMAC comes from the official ranking of "AI meets beauty"~\cite{acmchallenge}.
Moreover, to demonstrate the effectiveness of our dataset and for the sake of fairness, we train the backbone network of all methods on our dataset. As illustrated in Table \ref{compaire}, our dataset can significantly improve the performance and our method is still the best.

\begin{figure}[htbp]
	\centering
	\includegraphics[width=1.0\linewidth]{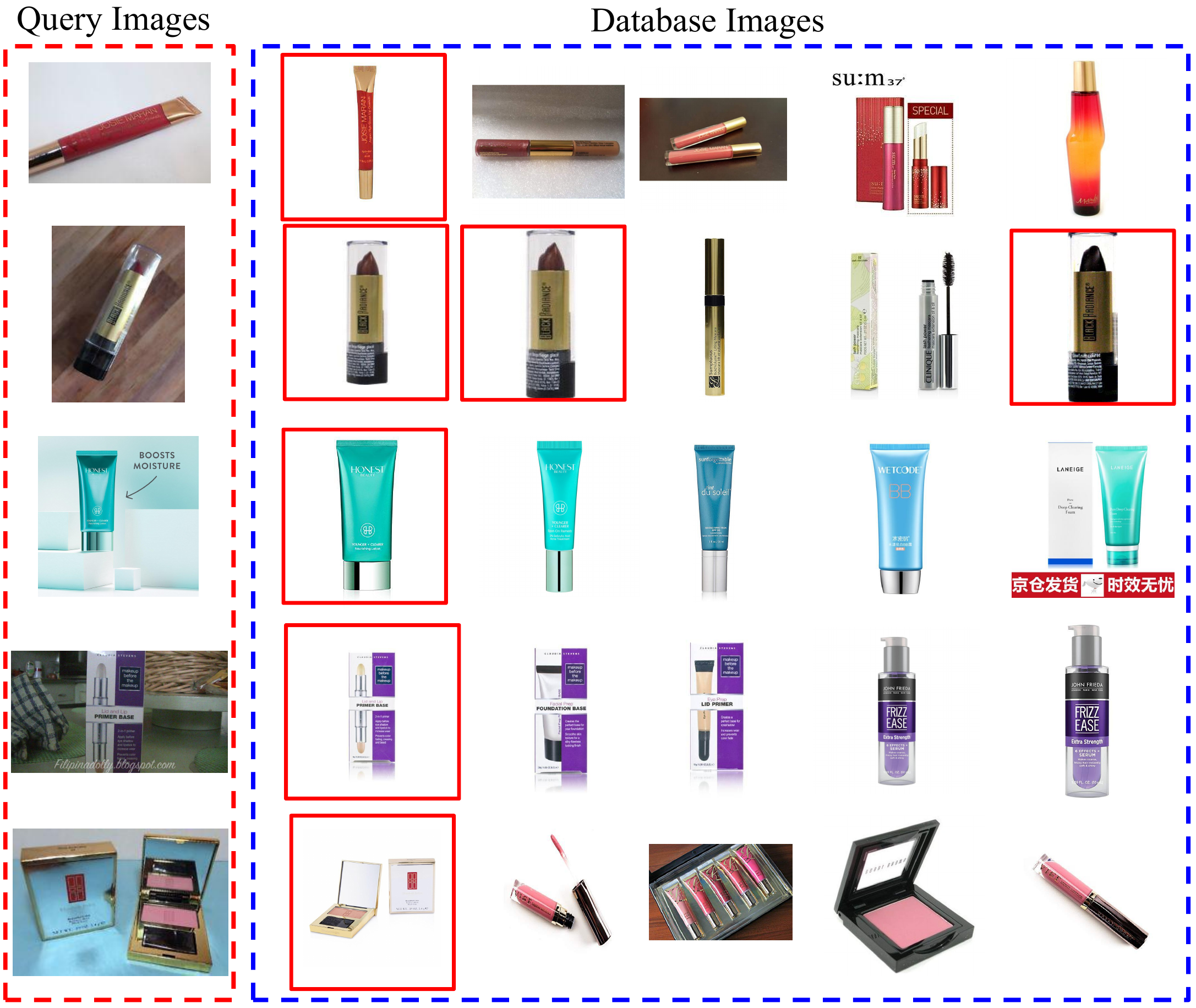}
	%
	\parbox[t]{.95\columnwidth}{
	}
	\caption{\label{fig:visuall}
		\noindent\textbf{Visualization of our retrieval results.} BPR is difficult for many similar interference options and data variations, but VM-Net can effectively find the same product under these interference factors.}
\end{figure}
\begin{figure}[t]
	\centering
	\includegraphics[width=1.0\linewidth]{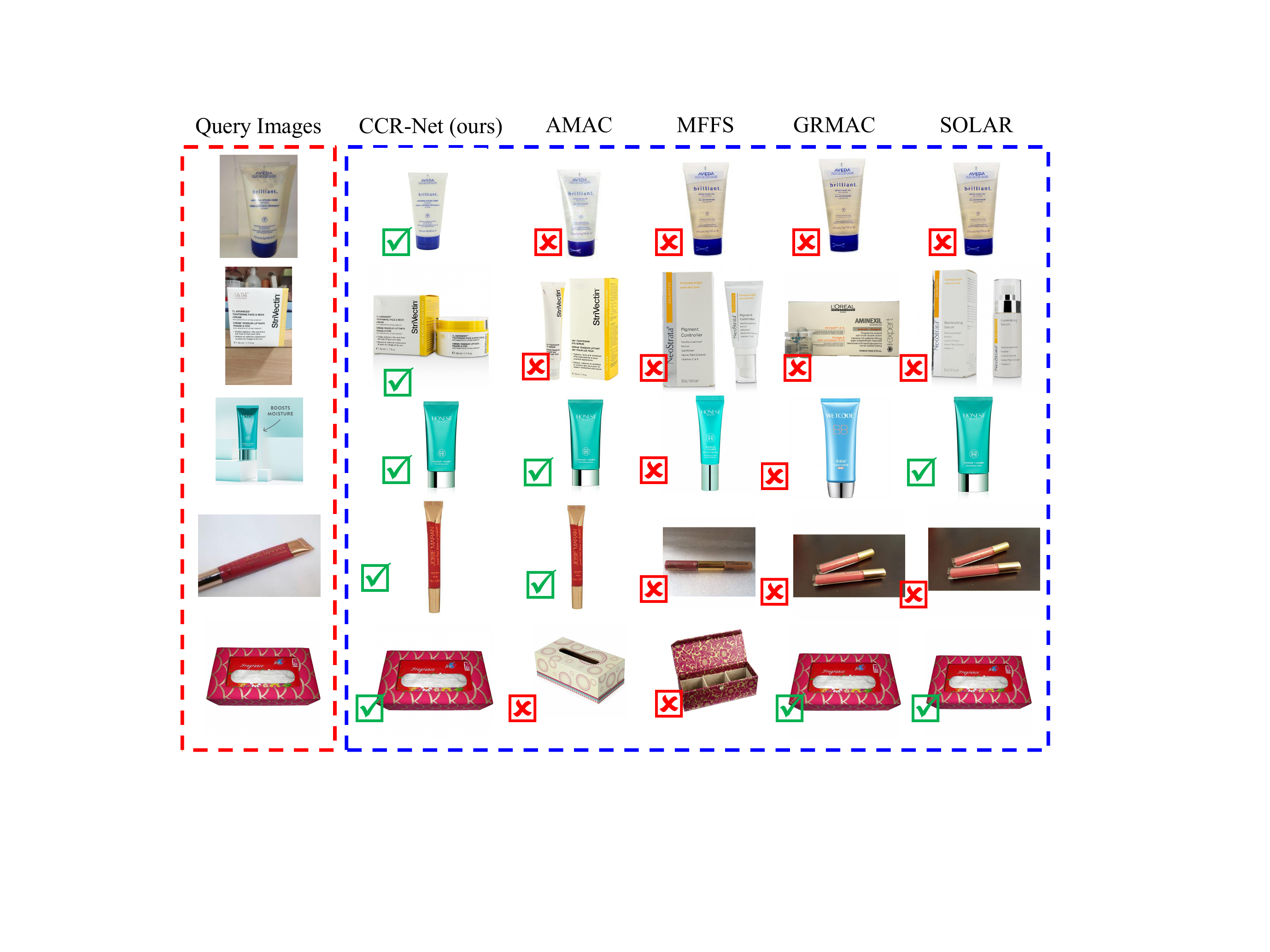}
	%
	\parbox[t]{.9\columnwidth}{
	}
	\caption{\label{fig:right}
		\noindent\textbf{VM-Net outperforms its competitors for challenging cases.} Given a query image, the top-1 retrieval results are illustrated. Other state-of-the-art methods find the similar images in most cases, while our VM-Net achieves exactly the same product's image.}
\end{figure}
\subsection{Visualization}
We randomly choose five samples from the results. As shown in Figure \ref{fig:visuall}, we show the top five most similar images selected by VM-Net in every sample and we mark the same product with a red box. It is observed that BPR is quite difficult to obtain satisfying results because the query images are full of data variations and database images which do not contain the target product may be very similar to the query image. Our method can pick out the exact images under these interference factors.

We list five typical query images with data variations from real world, and illustrate the top-1 retrieval results of state-of-the-arts and our VM-Net in Figure \ref{fig:right}. Other methods may search the most similar images, while our VM-Net achieves exactly the same product's image.

\section{Conclusion}
We have raised the intriguing question that whether the broadly extracted different types of image features are beneficial to 1) suppress data variations of real-world captured images, and 2) distinguish one image from others in the dataset that have very similar but intrinsically different beauty products in them, therefore leading to an enhanced capability of BPR. The answer is YES. That is, we propose an end-to-end content-based beauty product retrieval neural network, named VM-Net. VM-Net skillfully uses the variable attention mechanism and integrates features of various levels including low-level features (saliency features), middle-convolution-layer features, distinct features and detailed features. It is worth mentioning that saliency features and distinct features can effectively counter the data variations, while middle-convolution-layer features and detailed features can distinguish the similar products and the target product accurately. Extensive results show that VM-Net  outperforms the state-of-the-art methods clearly in terms of retrieval accuracy. After training on our own collected dataset, in which the proportion of beauty products is less than 5\%, VM-Net achieves the best results among all the methods on Perfect-500k. The outstanding performance exhibits a potentially wide applicability of VM-Net. In addition, VM-Net can be applied to the similar fields such as landmark retrieval, sketch retrieval, etc. Also, VM-Net can be applied to both the beauty product identification and beauty product classification tasks with minor modifications. For example, we may replace the module for calculating the similarity of the model with predicting the probability score and calculate the loss function.


\bibliographystyle{ACM-Reference-Format}
\bibliography{main}


\begin{thebibliography}{30}


\ifx \showCODEN    \undefined \def \showCODEN     #1{\unskip}     \fi
\ifx \showDOI      \undefined \def \showDOI       #1{#1}\fi
\ifx \showISBNx    \undefined \def \showISBNx     #1{\unskip}     \fi
\ifx \showISBNxiii \undefined \def \showISBNxiii  #1{\unskip}     \fi
\ifx \showISSN     \undefined \def \showISSN      #1{\unskip}     \fi
\ifx \showLCCN     \undefined \def \showLCCN      #1{\unskip}     \fi
\ifx \shownote     \undefined \def \shownote      #1{#1}          \fi
\ifx \showarticletitle \undefined \def \showarticletitle #1{#1}   \fi
\ifx \showURL      \undefined \def \showURL       {\relax}        \fi
\providecommand\bibfield[2]{#2}
\providecommand\bibinfo[2]{#2}
\providecommand\natexlab[1]{#1}
\providecommand\showeprint[2][]{arXiv:#2}

\bibitem[\protect\citeauthoryear{??}{acm}{2020}]%
        {acmchallenge}
 \bibinfo{year}{2020}\natexlab{}.
\newblock \showarticletitle{https://challenge2020.perfectcorp.com/}.
\newblock


\bibitem[\protect\citeauthoryear{??}{tao}{2020}]%
        {taobao}
 \bibinfo{year}{2020}\natexlab{}.
\newblock
  \showarticletitle{https://tianchi.aliyun.com/competition/entrance/231772/introduction}.
\newblock


\bibitem[\protect\citeauthoryear{Auli and Lopez}{Auli and Lopez}{2011}]%
        {DBLP:conf/emnlp/AuliL11}
\bibfield{author}{\bibinfo{person}{Michael Auli} {and} \bibinfo{person}{Adam
  Lopez}.} \bibinfo{year}{2011}\natexlab{}.
\newblock \showarticletitle{Training a Log-Linear Parser with Loss Functions
  via Softmax-Margin}. In \bibinfo{booktitle}{\emph{Proceedings of the 2011
  Conference on Empirical Methods in Natural Language Processing}}.
  \bibinfo{publisher}{{ACL}}, \bibinfo{pages}{333--343}.
\newblock


\bibitem[\protect\citeauthoryear{Brattoli, Roth, and Ommer}{Brattoli
  et~al\mbox{.}}{2019}]%
        {DBLP:conf/iccv/BrattoliRO19}
\bibfield{author}{\bibinfo{person}{Biagio Brattoli}, \bibinfo{person}{Karsten
  Roth}, {and} \bibinfo{person}{Bj{\"{o}}rn Ommer}.}
  \bibinfo{year}{2019}\natexlab{}.
\newblock \showarticletitle{{MIC:} Mining Interclass Characteristics for
  Improved Metric Learning}. In \bibinfo{booktitle}{\emph{2019 {IEEE/CVF}
  International Conference on Computer Vision}}. \bibinfo{pages}{7999--8008}.
\newblock


\bibitem[\protect\citeauthoryear{Brown, Xie, Kalogeiton, and Zisserman}{Brown
  et~al\mbox{.}}{2020}]%
        {DBLP:conf/eccv/BrownXKZ20}
\bibfield{author}{\bibinfo{person}{Andrew Brown}, \bibinfo{person}{Weidi Xie},
  \bibinfo{person}{Vicky Kalogeiton}, {and} \bibinfo{person}{Andrew
  Zisserman}.} \bibinfo{year}{2020}\natexlab{}.
\newblock \showarticletitle{Smooth-AP: Smoothing the Path Towards Large-Scale
  Image Retrieval}. In \bibinfo{booktitle}{\emph{Computer Vision - {ECCV} 2020
  - 16th European Conference}} \emph{(\bibinfo{series}{Lecture Notes in
  Computer Science}, Vol.~\bibinfo{volume}{12354})}. \bibinfo{pages}{677--694}.
\newblock


\bibitem[\protect\citeauthoryear{Datta, Joshi, Li, and Wang}{Datta
  et~al\mbox{.}}{2008}]%
        {DBLP:journals/csur/DattaJLW08}
\bibfield{author}{\bibinfo{person}{Ritendra Datta}, \bibinfo{person}{Dhiraj
  Joshi}, \bibinfo{person}{Jia Li}, {and} \bibinfo{person}{James~Ze Wang}.}
  \bibinfo{year}{2008}\natexlab{}.
\newblock \showarticletitle{Image retrieval: Ideas, influences, and trends of
  the new age}.
\newblock \bibinfo{journal}{\emph{{ACM} Comput. Surv.}} \bibinfo{volume}{40},
  \bibinfo{number}{2} (\bibinfo{year}{2008}), \bibinfo{pages}{5:1--5:60}.
\newblock
\urldef\tempurl%
\url{https://doi.org/10.1145/1348246.1348248}
\showDOI{\tempurl}


\bibitem[\protect\citeauthoryear{Deng, Dong, Socher, Li, Li, and Li}{Deng
  et~al\mbox{.}}{2009}]%
        {DBLP:conf/cvpr/DengDSLL009}
\bibfield{author}{\bibinfo{person}{Jia Deng}, \bibinfo{person}{Wei Dong},
  \bibinfo{person}{Richard Socher}, \bibinfo{person}{Li{-}Jia Li},
  \bibinfo{person}{Kai Li}, {and} \bibinfo{person}{Fei{-}Fei Li}.}
  \bibinfo{year}{2009}\natexlab{}.
\newblock \showarticletitle{ImageNet: {A} large-scale hierarchical image
  database}. In \bibinfo{booktitle}{\emph{2009 {IEEE} Computer Society
  Conference on Computer Vision and Pattern Recognition}}.
  \bibinfo{pages}{248--255}.
\newblock


\bibitem[\protect\citeauthoryear{Fang, Zhou, Roy, Petersson, and Harandi}{Fang
  et~al\mbox{.}}{2019}]%
        {DBLP:conf/iccv/FangZRPH19}
\bibfield{author}{\bibinfo{person}{Pengfei Fang}, \bibinfo{person}{Jieming
  Zhou}, \bibinfo{person}{Soumava~Kumar Roy}, \bibinfo{person}{Lars Petersson},
  {and} \bibinfo{person}{Mehrtash Harandi}.} \bibinfo{year}{2019}\natexlab{}.
\newblock \showarticletitle{Bilinear Attention Networks for Person Retrieval}.
  In \bibinfo{booktitle}{\emph{2019 {IEEE/CVF} International Conference on
  Computer Vision}}. \bibinfo{pages}{8029--8038}.
\newblock


\bibitem[\protect\citeauthoryear{Ge, Wang, Zhu, Zhao, and Li}{Ge
  et~al\mbox{.}}{2020}]%
        {DBLP:conf/eccv/GeW00L20}
\bibfield{author}{\bibinfo{person}{Yixiao Ge}, \bibinfo{person}{Haibo Wang},
  \bibinfo{person}{Feng Zhu}, \bibinfo{person}{Rui Zhao}, {and}
  \bibinfo{person}{Hongsheng Li}.} \bibinfo{year}{2020}\natexlab{}.
\newblock \showarticletitle{Self-supervising Fine-Grained Region Similarities
  for Large-Scale Image Localization}. In \bibinfo{booktitle}{\emph{Computer
  Vision - {ECCV} 2020 - 16th European Conference}}
  \emph{(\bibinfo{series}{Lecture Notes in Computer Science},
  Vol.~\bibinfo{volume}{12349})}. \bibinfo{pages}{369--386}.
\newblock


\bibitem[\protect\citeauthoryear{Jang and Cho}{Jang and Cho}{2020}]%
        {DBLP:conf/cvpr/JangC20}
\bibfield{author}{\bibinfo{person}{Young~Kyun Jang} {and}
  \bibinfo{person}{Nam~Ik Cho}.} \bibinfo{year}{2020}\natexlab{}.
\newblock \showarticletitle{Generalized Product Quantization Network for
  Semi-Supervised Image Retrieval}. In \bibinfo{booktitle}{\emph{2020
  {IEEE/CVF} Conference on Computer Vision and Pattern Recognition}}.
  \bibinfo{pages}{3417--3426}.
\newblock


\bibitem[\protect\citeauthoryear{Lee, Seong, Lee, and Kim}{Lee
  et~al\mbox{.}}{2022}]%
        {DBLP:journals/corr/abs-2204-01458}
\bibfield{author}{\bibinfo{person}{Seongwon Lee}, \bibinfo{person}{Hongje
  Seong}, \bibinfo{person}{Suhyeon Lee}, {and} \bibinfo{person}{Euntai Kim}.}
  \bibinfo{year}{2022}\natexlab{}.
\newblock \showarticletitle{Correlation Verification for Image Retrieval}.
\newblock \bibinfo{journal}{\emph{CoRR}}  \bibinfo{volume}{abs/2204.01458}
  (\bibinfo{year}{2022}).
\newblock


\bibitem[\protect\citeauthoryear{Lin, Doll{\'{a}}r, Girshick, He, Hariharan,
  and Belongie}{Lin et~al\mbox{.}}{2017}]%
        {cvpr/LinDGHHB17}
\bibfield{author}{\bibinfo{person}{Tsung{-}Yi Lin}, \bibinfo{person}{Piotr
  Doll{\'{a}}r}, \bibinfo{person}{Ross~B. Girshick}, \bibinfo{person}{Kaiming
  He}, \bibinfo{person}{Bharath Hariharan}, {and} \bibinfo{person}{Serge~J.
  Belongie}.} \bibinfo{year}{2017}\natexlab{}.
\newblock \showarticletitle{Feature Pyramid Networks for Object Detection}. In
  \bibinfo{booktitle}{\emph{2017 {IEEE} Conference on Computer Vision and
  Pattern Recognition}}. \bibinfo{pages}{936--944}.
\newblock


\bibitem[\protect\citeauthoryear{Lin, Xie, Kang, Yang, Liu, and Li}{Lin
  et~al\mbox{.}}{2019}]%
        {2019Beautyi}
\bibfield{author}{\bibinfo{person}{Zehang Lin}, \bibinfo{person}{Haoran Xie},
  \bibinfo{person}{Peipei Kang}, \bibinfo{person}{Zhenguo Yang},
  \bibinfo{person}{Wenyin Liu}, {and} \bibinfo{person}{Qing Li}.}
  \bibinfo{year}{2019}\natexlab{}.
\newblock \showarticletitle{Cross-domain Beauty Item Retrieval via Unsupervised
  Embedding Learning}. In \bibinfo{booktitle}{\emph{Proceedings of the 27th
  {ACM} International Conference on Multimedia}}. \bibinfo{pages}{2543--2547}.
\newblock


\bibitem[\protect\citeauthoryear{Ng, Balntas, Tian, and Mikolajczyk}{Ng
  et~al\mbox{.}}{[n.d.]}]%
        {DBLP:conf/eccv/NgBTM20}
\bibfield{author}{\bibinfo{person}{Tony Ng}, \bibinfo{person}{Vassileios
  Balntas}, \bibinfo{person}{Yurun Tian}, {and} \bibinfo{person}{Krystian
  Mikolajczyk}.} \bibinfo{year}{[n.d.]}\natexlab{}.
\newblock \showarticletitle{{SOLAR:} Second-Order Loss and Attention for Image
  Retrieval}. In \bibinfo{booktitle}{\emph{Computer Vision - {ECCV} 2020 - 16th
  European Conference}} \emph{(\bibinfo{series}{Lecture Notes in Computer
  Science}, Vol.~\bibinfo{volume}{12370})}. \bibinfo{pages}{253--270}.
\newblock


\bibitem[\protect\citeauthoryear{Redmon and Farhadi}{Redmon and
  Farhadi}{2017}]%
        {DBLP:conf/cvpr/RedmonF17}
\bibfield{author}{\bibinfo{person}{Joseph Redmon} {and} \bibinfo{person}{Ali
  Farhadi}.} \bibinfo{year}{2017}\natexlab{}.
\newblock \showarticletitle{{YOLO9000:} Better, Faster, Stronger}. In
  \bibinfo{booktitle}{\emph{2017 {IEEE} Conference on Computer Vision and
  Pattern Recognition}}. \bibinfo{pages}{6517--6525}.
\newblock


\bibitem[\protect\citeauthoryear{Redmon and Farhadi}{Redmon and
  Farhadi}{2018}]%
        {DBLP:journals/corr/abs-1804-02767}
\bibfield{author}{\bibinfo{person}{Joseph Redmon} {and} \bibinfo{person}{Ali
  Farhadi}.} \bibinfo{year}{2018}\natexlab{}.
\newblock \showarticletitle{YOLOv3: An Incremental Improvement}.
\newblock \bibinfo{journal}{\emph{CoRR}}  \bibinfo{volume}{abs/1804.02767}
  (\bibinfo{year}{2018}).
\newblock


\bibitem[\protect\citeauthoryear{Revaud, Almaz{\'{a}}n, Rezende, and
  de~Souza}{Revaud et~al\mbox{.}}{[n.d.]}]%
        {DBLP:conf/iccv/RevaudARS19}
\bibfield{author}{\bibinfo{person}{J{\'{e}}r{\^{o}}me Revaud},
  \bibinfo{person}{Jon Almaz{\'{a}}n}, \bibinfo{person}{Rafael~S. Rezende},
  {and} \bibinfo{person}{C{\'{e}}sar~Roberto de Souza}.}
  \bibinfo{year}{[n.d.]}\natexlab{}.
\newblock \showarticletitle{Learning With Average Precision: Training Image
  Retrieval With a Listwise Loss}. In \bibinfo{booktitle}{\emph{2019 {IEEE/CVF}
  International Conference on Computer Vision}}. \bibinfo{pages}{5106--5115}.
\newblock


\bibitem[\protect\citeauthoryear{Sun, Cheng, Zhang, Zhang, Zheng, Wang, and
  Wei}{Sun et~al\mbox{.}}{[n.d.]}]%
        {DBLP:conf/cvpr/SunCZZZWW20}
\bibfield{author}{\bibinfo{person}{Yifan Sun}, \bibinfo{person}{Changmao
  Cheng}, \bibinfo{person}{Yuhan Zhang}, \bibinfo{person}{Chi Zhang},
  \bibinfo{person}{Liang Zheng}, \bibinfo{person}{Zhongdao Wang}, {and}
  \bibinfo{person}{Yichen Wei}.} \bibinfo{year}{[n.d.]}\natexlab{}.
\newblock \showarticletitle{Circle Loss: {A} Unified Perspective of Pair
  Similarity Optimization}. In \bibinfo{booktitle}{\emph{2020 {IEEE/CVF}
  Conference on Computer Vision and Pattern Recognition}}.
  \bibinfo{pages}{6397--6406}.
\newblock


\bibitem[\protect\citeauthoryear{Tolias, Sicre, and J{\'{e}}gou}{Tolias
  et~al\mbox{.}}{2016}]%
        {2016RMAC}
\bibfield{author}{\bibinfo{person}{Giorgos Tolias}, \bibinfo{person}{Ronan
  Sicre}, {and} \bibinfo{person}{Herv{\'{e}} J{\'{e}}gou}.}
  \bibinfo{year}{2016}\natexlab{}.
\newblock \showarticletitle{Particular object retrieval with integral
  max-pooling of {CNN} activations}. In \bibinfo{booktitle}{\emph{4th
  International Conference on Learning Representations}}.
\newblock


\bibitem[\protect\citeauthoryear{Wang, Zhu, Xu, and Cao}{Wang
  et~al\mbox{.}}{2019}]%
        {2019The}
\bibfield{author}{\bibinfo{person}{Jiawei Wang}, \bibinfo{person}{Shuai Zhu},
  \bibinfo{person}{Jiao Xu}, {and} \bibinfo{person}{Da Cao}.}
  \bibinfo{year}{2019}\natexlab{}.
\newblock \showarticletitle{The Retrieval of the Beautiful: Self-Supervised
  Salient Object Detection for Beauty Product Retrieval}. In
  \bibinfo{booktitle}{\emph{Proceedings of the 27th {ACM} International
  Conference on Multimedia}}. \bibinfo{pages}{2548--2552}.
\newblock


\bibitem[\protect\citeauthoryear{Wang, Liu, Lin, Yang, and Li}{Wang
  et~al\mbox{.}}{2020}]%
        {2020Multi}
\bibfield{author}{\bibinfo{person}{Zhihui Wang}, \bibinfo{person}{Xing Liu},
  \bibinfo{person}{Jiawen Lin}, \bibinfo{person}{Caifei Yang}, {and}
  \bibinfo{person}{Haojie Li}.} \bibinfo{year}{2020}\natexlab{}.
\newblock \showarticletitle{Multi-attention based cross-domain beauty product
  image retrieval}.
\newblock \bibinfo{journal}{\emph{Sci. China Inf. Sci.}} \bibinfo{volume}{63},
  \bibinfo{number}{2} (\bibinfo{year}{2020}), \bibinfo{pages}{120112}.
\newblock


\bibitem[\protect\citeauthoryear{Wei, Wang, Wu, Su, Huang, and Tian}{Wei
  et~al\mbox{.}}{2020}]%
        {DBLP:conf/cvpr/WeiWWSH020}
\bibfield{author}{\bibinfo{person}{Jun Wei}, \bibinfo{person}{Shuhui Wang},
  \bibinfo{person}{Zhe Wu}, \bibinfo{person}{Chi Su}, \bibinfo{person}{Qingming
  Huang}, {and} \bibinfo{person}{Qi Tian}.} \bibinfo{year}{2020}\natexlab{}.
\newblock \showarticletitle{Label Decoupling Framework for Salient Object
  Detection}. In \bibinfo{booktitle}{\emph{2020 {IEEE/CVF} Conference on
  Computer Vision and Pattern Recognition}}. \bibinfo{publisher}{{IEEE}},
  \bibinfo{pages}{13022--13031}.
\newblock


\bibitem[\protect\citeauthoryear{Weinzaepfel, Lucas, Larlus, and
  Kalantidis}{Weinzaepfel et~al\mbox{.}}{2022}]%
        {DBLP:journals/corr/abs-2201-13182}
\bibfield{author}{\bibinfo{person}{Philippe Weinzaepfel},
  \bibinfo{person}{Thomas Lucas}, \bibinfo{person}{Diane Larlus}, {and}
  \bibinfo{person}{Yannis Kalantidis}.} \bibinfo{year}{2022}\natexlab{}.
\newblock \showarticletitle{Learning Super-Features for Image Retrieval}.
\newblock \bibinfo{journal}{\emph{CoRR}}  \bibinfo{volume}{abs/2201.13182}
  (\bibinfo{year}{2022}).
\newblock


\bibitem[\protect\citeauthoryear{Xu and Liu}{Xu and Liu}{2020}]%
        {2007A}
\bibfield{author}{\bibinfo{person}{Nianli Xu} {and} \bibinfo{person}{Fengying
  Liu}.} \bibinfo{year}{2020}\natexlab{}.
\newblock \showarticletitle{Application of image content feature retrieval
  based on deep learning in sports public industry}.
\newblock \bibinfo{journal}{\emph{J. Intell. Fuzzy Syst.}}
  \bibinfo{volume}{39}, \bibinfo{number}{2} (\bibinfo{year}{2020}),
  \bibinfo{pages}{1867--1877}.
\newblock


\bibitem[\protect\citeauthoryear{Xuan, Stylianou, Liu, and Pless}{Xuan
  et~al\mbox{.}}{[n.d.]}]%
        {DBLP:conf/eccv/XuanSLP20}
\bibfield{author}{\bibinfo{person}{Hong Xuan}, \bibinfo{person}{Abby
  Stylianou}, \bibinfo{person}{Xiaotong Liu}, {and} \bibinfo{person}{Robert
  Pless}.} \bibinfo{year}{[n.d.]}\natexlab{}.
\newblock \showarticletitle{Hard Negative Examples are Hard, but Useful}. In
  \bibinfo{booktitle}{\emph{Computer Vision - {ECCV} 2020 - 16th European
  Conference}} \emph{(\bibinfo{series}{Lecture Notes in Computer Science},
  Vol.~\bibinfo{volume}{12359})}. \bibinfo{pages}{126--142}.
\newblock


\bibitem[\protect\citeauthoryear{Yan, Lin, Deng, Lei, and Xu}{Yan
  et~al\mbox{.}}{[n.d.]}]%
        {DBLP:conf/mm/YanLDLX20}
\bibfield{author}{\bibinfo{person}{Runming Yan}, \bibinfo{person}{Yongchun
  Lin}, \bibinfo{person}{Zhichao Deng}, \bibinfo{person}{Liang Lei}, {and}
  \bibinfo{person}{Chudong Xu}.} \bibinfo{year}{[n.d.]}\natexlab{}.
\newblock \showarticletitle{Multi-Feature Fusion Method Based on Salient Object
  Detection for Beauty Product Retrieval}. In \bibinfo{booktitle}{\emph{{MM}
  '20: The 28th {ACM} International Conference on Multimedia}}.
  \bibinfo{pages}{4713--4717}.
\newblock


\bibitem[\protect\citeauthoryear{Yu, Xie, Li, Xie, Hao, Gao, and Shuang}{Yu
  et~al\mbox{.}}{2020}]%
        {DBLP:conf/mm/YuXLXHGS20}
\bibfield{author}{\bibinfo{person}{Jun Yu}, \bibinfo{person}{Guochen Xie},
  \bibinfo{person}{Mengyan Li}, \bibinfo{person}{Haonian Xie},
  \bibinfo{person}{Xinlong Hao}, \bibinfo{person}{Fang Gao}, {and}
  \bibinfo{person}{Feng Shuang}.} \bibinfo{year}{2020}\natexlab{}.
\newblock \showarticletitle{Attention Based Beauty Product Retrieval Using
  Global and Local Descriptors}. In \bibinfo{booktitle}{\emph{{MM} '20: The
  28th {ACM} International Conference on Multimedia}},
  \bibfield{editor}{\bibinfo{person}{Chang~Wen Chen}, \bibinfo{person}{Rita
  Cucchiara}, \bibinfo{person}{Xian{-}Sheng Hua}, \bibinfo{person}{Guo{-}Jun
  Qi}, \bibinfo{person}{Elisa Ricci}, \bibinfo{person}{Zhengyou Zhang}, {and}
  \bibinfo{person}{Roger Zimmermann}} (Eds.). \bibinfo{pages}{4708--4712}.
\newblock


\bibitem[\protect\citeauthoryear{Yu, Xie, Li, Xie, and Yu}{Yu
  et~al\mbox{.}}{2019}]%
        {2019Beauty}
\bibfield{author}{\bibinfo{person}{Jun Yu}, \bibinfo{person}{Guochen Xie},
  \bibinfo{person}{Mengyan Li}, \bibinfo{person}{Haonian Xie}, {and}
  \bibinfo{person}{Lingyun Yu}.} \bibinfo{year}{2019}\natexlab{}.
\newblock \showarticletitle{Beauty Product Retrieval Based on Regional Maximum
  Activation of Convolutions with Generalized Attention}. In
  \bibinfo{booktitle}{\emph{Proceedings of the 27th {ACM} International
  Conference on Multimedia}}. \bibinfo{pages}{2553--2557}.
\newblock


\bibitem[\protect\citeauthoryear{Zhang, Qu, He, Lu, and Gao}{Zhang
  et~al\mbox{.}}{2019}]%
        {2019Beaut}
\bibfield{author}{\bibinfo{person}{Yi Zhang}, \bibinfo{person}{Linzi Qu},
  \bibinfo{person}{Lihuo He}, \bibinfo{person}{Wen Lu}, {and}
  \bibinfo{person}{Xinbo Gao}.} \bibinfo{year}{2019}\natexlab{}.
\newblock \showarticletitle{Beauty Aware Network: An Unsupervised Method for
  Makeup Product Retrieval}. In \bibinfo{booktitle}{\emph{Proceedings of the
  27th {ACM} International Conference on Multimedia}}.
  \bibinfo{pages}{2558--2562}.
\newblock


\bibitem[\protect\citeauthoryear{Zheng, Yang, and Tian}{Zheng
  et~al\mbox{.}}{2018}]%
        {2018SIFT}
\bibfield{author}{\bibinfo{person}{Liang Zheng}, \bibinfo{person}{Yi Yang},
  {and} \bibinfo{person}{Qi Tian}.} \bibinfo{year}{2018}\natexlab{}.
\newblock \showarticletitle{SIFT Meets CNN: A Decade Survey of Instance
  Retrieval}.
\newblock \bibinfo{journal}{\emph{IEEE Transactions on Pattern Analysis and
  Machine Intelligence}} \bibinfo{number}{5} (\bibinfo{year}{2018}),
  \bibinfo{pages}{1224--1244}.
\newblock


\end{thebibliography}


\end{document}